\pdfoutput=1
\documentclass{article}

\usepackage{PRIMEarxiv}

\usepackage[utf8]{inputenc}
\usepackage[T1]{fontenc}
\usepackage{xcolor}
\usepackage[colorlinks=true,linkcolor=blue!60!black,citecolor=blue!60!black,urlcolor=blue!60!black]{hyperref}
\usepackage{url}
\usepackage{booktabs,array}
\usepackage{amsmath,amssymb,amsthm}
\usepackage{mathtools}
\usepackage{microtype}
\usepackage{fancyhdr}
\usepackage{graphicx}
\usepackage[numbers,sort&compress]{natbib}

\allowdisplaybreaks

\newtheorem{theorem}{Theorem}
\newtheorem{proposition}[theorem]{Proposition}
\theoremstyle{definition}
\newtheorem{remark}[theorem]{Remark}

\newcommand{\R}{\mathbb R}
\newcommand{\E}{\mathcal E}
\newcommand{\V}{\mathcal V}
\DeclarePairedDelimiter{\norm}{\lVert}{\rVert}
\DeclareMathOperator{\col}{col}
\DeclareMathOperator{\vecop}{vec}

\title{Efficient Modeling of Multi-Agent Systems via Structure-Aware Physics-Informed Learning}

\author{
  Yulun Wu \\
  Division of Decision and Control Systems \\
  KTH Royal Institute of Technology \\
  Stockholm, Sweden \\
  \texttt{yulunw@kth.se} \\
  \And
  Nicola Bastianello \\
  Department of Electronic \& Electrical Engineering \\
  University College London \\
  United Kingdom \\
  \texttt{n.bastianello@ucl.ac.uk} \\
  \AND
  Karl Henrik Johansson \\
  Division of Decision and Control Systems \\
  KTH Royal Institute of Technology \\
  Stockholm, Sweden \\
  \texttt{kallej@kth.se} \\
  \And
  Matthieu Barreau \\
  Division of Decision and Control Systems \\
  KTH Royal Institute of Technology \\
  Stockholm, Sweden \\
  \texttt{barreau@kth.se} \\
}

\begin{document}
\maketitle

\begin{abstract}
Modeling large multi-agent systems poses challenges for trajectory inference and parameter identification, as network size and coupling increase computational demands and partial observations require reconstructing unobserved states.
We propose MAS-PINN, a neural-network architecture with a tailored training algorithm for a physics-informed learning framework that leverages the network topology and the agents' similarity to reduce computational and memory costs of simulating and identifying multi-agent systems. Its architecture represents agent trajectories through shared temporal features and agent-specific coefficients. Training alternates between learning features, fitting coupled trajectories, and estimating physical parameters. A matrix-free update exploits local interactions to fit trajectories without forming large global matrices.
Experiments on Lorenz rings, Lotka--Volterra networks, and Rössler oscillators show improved trajectory and parameter accuracy over baselines, including inference in large systems and identification under partial observations or unknown topology, alongside substantial computational savings in coefficient fitting.
\end{abstract}


\section{Introduction}

Accurate models of multi-agent systems (MAS) are crucial for understanding and predicting collective behavior in applications such as cooperative robotics, power systems, and ecological communities \citep{olfati2007consensus,dorfler2014synchronization,allesina2012stability}. These models describe nonlinear interactions through coupled ordinary differential equations with specificities.
First, the agents share a nominal model of their own dynamics, with agent-specific variations. Second, the coupling is established over a typically fixed network. Finally, even if the coupled system is of large dimension, its global dynamical representation is sparse and structured.
Classical approaches to simulate and identify such systems combine mechanistic models with numerical
integration and parameter estimation, using nonlinear state-space
identification and multiple shooting
\citep{schon2011system,peifer2007parameter}.
However, incomplete knowledge of system dynamics and limited observations complicate model construction and calibration.
Learning-based approaches offer more flexible representations of
nonlinear relationships that can incorporate available physical
knowledge \citep{rackauckas2020universal} at the price of less rigorous convergence guarantees.

Among the learning frameworks, Physics-informed neural networks (PINNs) have been effective at combining observations with governing equations by incorporating physics residuals into the training objective~\citep{raissi2019physics,karniadakis2021physics}. Variations of PINNs have been developed for state reconstruction, inverse problems, control, and network identification \citep{raissi2020hidden,yang2021bpinns,antonelo2024control,barreau2021physics} in a large variety of applications. Nevertheless, they face multiple issues in a MAS setting~\citep{wang2022ntk,krishnapriyan2021failure,rathore2024challenges}.
Indeed, large multi-agent systems involve large state variables with memory issues
and coupled dynamical residuals which hinder training with high computational burden.
This motivates architectures and numerical methods that exploit
shared dynamics and coupling to improve computational efficiency.

Considering physics-informed approaches, \citet{yang2026multi} demonstrate that PINNs are appropriate for MAS modeling, observation, and control problems. However, they do not consider the memory and computational bottlenecks present in large MAS.
Existing approaches to address these challenges include staged training for network identification~\citep{mao2025hyperpinn}, bilevel optimization~\citep{zhang2026bilo}, and pretrained bases for repeated
inverse queries \citep{manor2026ip}.
The FALM framework \citep{wu2026alternating} proposes to separate representation
learning from output-coefficient fitting, combining
Fourier-enhanced features with nonlinear least-squares updates.
None of these methods apply directly to MAS where interactions between agents
require their trajectories to be fitted jointly, still exposing us to the curse of dimensionality.

\begin{figure}[htbp!]
  \centering
  \includegraphics[width= 1.01\linewidth]{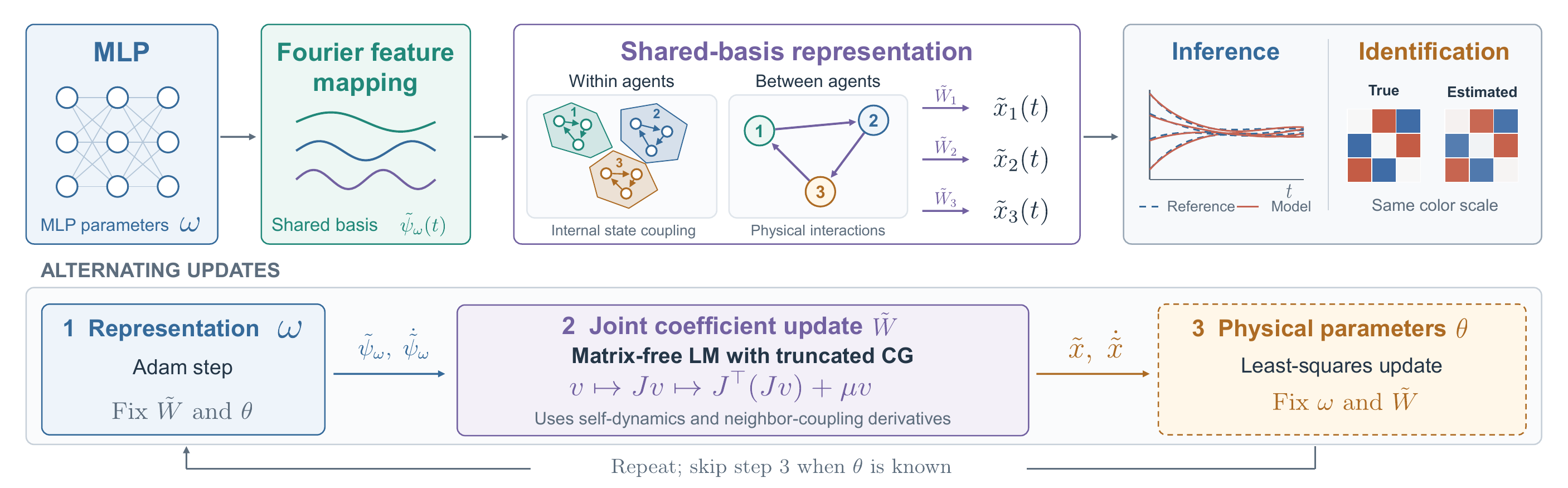}
  \caption{Architecture of MAS-PINN. The first row illustrates the shared basis and
agent-specific coefficients.
The second row shows alternating updates of the representation,
trajectory coefficients, and physical parameters, with the
last update used only for identification.}
  \label{fig:StructPINN}
\end{figure}

In this paper, we propose MAS-PINN, which combines a shared neural representation with training tailored to network sparsity for trajectory inference and parameter identification (Figure~\ref{fig:StructPINN}).
Our contributions are threefold: (i) we represent agent trajectories using a shared temporal basis
with agent-specific coefficients and develop an alternating
training scheme for inference with known physical parameters.
We then extend this to three-block alternating training for
parameter identification by introducing a least-squares
parameter update;
(ii) we develop a matrix-free Levenberg--Marquardt (LM) update
with truncated conjugate gradients that exploits local dynamics
and neighbor couplings without assembling the full Jacobian,
while retaining joint coefficient updates across agents;
and
(iii) we evaluate reconstruction accuracy and parameter
recovery against existing methods, and assess computational
cost and scaling with network size.

The remainder of this paper is organized as follows. Section~\ref{sec:preliminaries} introduces the preliminaries. The problem formulation is described in Section~\ref{sec:problem-statement}. Section~\ref{sec:method} presents the algorithm in a matrix-free implementation. Numerical experiments are reported in Section~\ref{sec:experiments}, and Section~\ref{sec:conclusion} concludes.

\section{Preliminaries}
\label{sec:preliminaries}

\subsection{Multi-agent systems}
\label{sec:network-model}

We consider a MAS of $N$ agents interacting over a directed graph $(\V,\E)$, where $\V=\{1,\ldots,N\}$ indexes the agents and an edge
$(j,i)\in\E$ indicates that agent $j$ influences agent $i$.
The set $\mathcal N_i=\{j:(j,i)\in\E\}$ denotes agent $i$'s neighbors with which it might communicate.
Agent $i$ has state $x_i(t)\in\R^d$ over $[0,T]$.

The dynamics of the agent is separated into two parts. The local dynamics $f(\cdot;\alpha_i)$ describe the evolution
of the $d$ state components of agent $i$. The coupling term $g(\cdot,\cdot;\beta_{ij})$ describes the
influence of a neighboring agent $j$ on agent $i$. Combining the two terms, we assume the following dynamics:
\begin{equation}
\begin{aligned}
\label{eq:agent-dynamics}
 \dot x_i &= f(x_i;\alpha_i)+\sum_{j\in\mathcal N_i} g(x_i,x_j;\beta_{ij}),
  \qquad i\in\V,
\end{aligned}
\end{equation}
where $f:\R^d\times\R^{q_\alpha}\to\R^d$ and
$g:\R^d\times\R^d\times\R^{q_\beta}\to\R^d$
are prescribed maps smooth in their state arguments.
The local parameters $\alpha_i\in\R^{q_\alpha}$ may differ slightly across agents, and $\beta_{ij}\in\R^{q_\beta}$ governs agent $j$'s influence on agent $i$.
The interaction structure and parameters are fixed over $[0,T]$.

Let $x=\col(x_1,\ldots,x_N)\in\R^n$, with $n=Nd$, and define
$\alpha=\col(\alpha_1,\ldots,\alpha_N)$.
Let $\beta$ collect the vectors $\beta_{ij}$ over all edges $(j,i)\in\E$ in a fixed order.
The stacked dynamics take the form
\begin{equation}
\begin{aligned}
\label{eq:stacked}
  \dot x&=\mathcal F(x;\alpha)+\mathcal G(x;\beta),
\end{aligned}
\end{equation}
where
\[
 \mathcal F(x;\alpha)
  =\col_{i=1}^{N}\bigl(f(x_i;\alpha_i)\bigr),
  \qquad
  \mathcal G(x;\beta)
  =\col_{i=1}^{N}
  \Bigl(
    \sum_{j\in\mathcal N_i}g(x_i,x_j;\beta_{ij})
  \Bigr).
\]
Thus, $\mathcal F$ collects the local dynamics, whereas $\mathcal G$
depends on the MAS topology.

\subsection{Physics-informed neural networks and FALM-PINN}
\label{sec:representation}
A PINN approximates the solution
$t \mapsto x(t)\in\R^n$ of \eqref{eq:stacked} by a multilayer perceptron
   $ x_\omega(t)=W^\top z_\omega(t)+b \in \R^n,$
where $z_\omega(t)\in\R^p$ is the output of the last hidden layer, $\omega$ collects the hidden layer parameters, and $(W,b) \in \R^{p\times n} \times \R^n$ is the linear output layer.
Training minimizes the following objective:
\begin{equation}
\label{eq:pinn-loss}
  \mathcal L(\omega,W,b)
  =
  \frac{1}{N_{\mathrm d}}\sum_{m=1}^{N_{\mathrm d}}
  \bigl\|r_{d}(t_m)\bigr\|^2
  +
  \frac{\lambda_f}{N_f}\sum_{\ell=1}^{N_f}
  \bigl\|r_\omega(\tau_\ell)\bigr\|^2,
\end{equation}
where $N_{\mathrm d}$ is the number of data points and $r_d$
is the data residual.
For a prescribed initial state, $N_{\mathrm d}=1$, and $r_d(0)=x_\omega(0)-x(0)$.
The sequence $\{\tau_\ell\}_\ell$ consists of $N_f$ collocation points in $[0, T]$,
$\lambda_f>0$ is the physics weight, and the physics
residual is given by
\begin{equation}
\label{eq:pinn-residual}
  r_\omega(t)=\dot{x}_\omega(t)
   -\mathcal F\bigl(x_\omega(t);\alpha\bigr)
   -\mathcal G\bigl(x_\omega(t);\beta\bigr).
\end{equation}
Standard PINN jointly updates $\omega$ and $(W,b)$  by gradient descent with specified hyperparameters.

\citet{wu2026alternating} proposed FALM-PINN, which treats the hidden layers and output layer as two blocks. It applies a random Fourier
feature mapping \citep{tancik2020fourier} to the last hidden layer to generate a wider range of independent features proven to map the solution space \citep{wu2025iterative}:
\begin{equation}
\label{eq:fourier-basis}
  \psi_\omega(t)=D_f^{-1/2}
  \begin{bmatrix}
    \cos\bigl(2\pi R^\top z_\omega(t)\bigr)\\
    \sin\bigl(2\pi R^\top z_\omega(t)\bigr)
  \end{bmatrix}
  \in\R^{2D_f},
\end{equation}
where the entries of $R\in\R^{p\times D_f}$ are drawn independently from $\mathcal N(0,\sigma^2)$, with bandwidth
$\sigma>0$ governing the frequency range of the features.
The output layer then acts on these features,
\begin{equation}
\label{eq:network-output}
\begin{aligned}
  \tilde{x}_\omega(t)
  =W^\top\psi_\omega(t)+b
   =\tilde{W}^\top\tilde{\psi}_\omega(t), \quad
  \tilde{\psi}_\omega(t)
  =\col\bigl(\psi_\omega(t),1\bigr)\in\R^P,
  \quad
  \tilde{W}=\begin{bmatrix}W\\b^\top\end{bmatrix}\in\R^{P\times n}.
\end{aligned}
\end{equation}
Here $W\in\R^{2D_f\times n}$, $b\in\R^n$, and $P=2D_f+1$.
The constant entry of the Fourier-enhanced basis $\tilde{\psi}_\omega$ incorporates the bias.
FALM-PINN alternates gradient updates of $\omega$ and
Levenberg--Marquardt (LM) steps
\citep{levenberg1944method,marquardt1963algorithm}
for the nonlinear least-squares problem in $\tilde{W}$,
keeping the other block fixed, therefore decoupling the nominal basis training in $\omega$ to the feature selection with $\tilde W$ and proven to converge to a local minimum.

\section{Problem statement}
\label{sec:problem-statement}
In this paper, we consider two problems: inference and identification for the MAS~\eqref{eq:stacked}.

\paragraph{Inference problem}
The parameters $\theta$ and the initial state $x(0)$ are given. The objective is to approximate the trajectory $x(t)$ over $[0,T]$ without computing the solution to~\eqref{eq:stacked} directly. We investigate how to adapt the FALM-PINN methodology to ensure that building a surrogate for \eqref{eq:stacked} is memory- and computationally efficient while preserving high accuracy.

\paragraph{Identification problem}
Assume the candidate graph $(\V,\E)$ and the functional forms of $f$ and $g$ are known, whereas the parameter vector $\theta=\col(\alpha,\beta)$ is unknown. We observe a single trajectory at sampling times:
\begin{equation}
\label{eq:observations}
  y_m= \mathcal{C} x(t_m)+\varepsilon_m,
  \qquad m=1,\ldots,N_{\mathrm{d}},
\end{equation}
where $\{t_m\}$ represents the sampling times, $\mathcal{C} \in \mathbb{R}^{n_s \times n}$ is the known observation matrix, and $\varepsilon_m\in\R^{n_s}$ denotes zero-mean measurement noise. Full-agent observations correspond to $\mathcal{C}=I_n$. For partial observations, $\mathcal{C}$ might select the complete states of agents in $\V_{\mathrm{o}}\subsetneq\V$, with $n_s=d|\V_{\mathrm{o}}|$, while the remaining agents $\V_{\mathrm{h}}=\V\setminus\V_{\mathrm{o}}$ are unobserved.

We want to jointly estimate $\theta$ and reconstruct
the complete trajectory $x(t)$ over $[0,T]$, including unobserved components, by fitting the observations subject to the governing dynamics \eqref{eq:stacked}. The data residual in that case is $r_d(t_m)=\mathcal Cx_\omega(t_m)-y_m$.

\begin{remark}
    We assume that the observations locally uniquely determine both the physical parameters $\theta$ and the initial state $x(0)$. This constitutes a joint assumption of local identifiability and observability for the MAS under consideration, which may be difficult to assess in practice.
\end{remark}

These two problems will be dealt with jointly.

\section{The MAS-PINN framework}
\label{sec:method}
MAS-PINN represents all agent trajectories using a shared
temporal basis and agent-specific output coefficients
(Figure~\ref{fig:StructPINN}) to resolve the memory problem.
Building on the alternating training scheme of FALM-PINN
\citep{wu2026alternating}, we develop a framework for joint state and parameter identification, alternating between updates to the basis, trajectory coefficients, and physical parameters. The coefficient update uses a matrix-free Levenberg--Marquardt (LM) solver that incorporates the shared basis and local dependencies in the agent dynamics. Therefore, without forming the global coefficient Jacobian or its normal matrix, the training algorithm is memory efficient if the network is sparse. We then recover inference with prescribed physical parameters as a special case of this training framework.

\subsection{Joint state and parameter identification}
\label{sec:parameter-update}
\label{sec:structured-update}

In our setting, both the trajectories and physical
parameters $\theta$ are unknown. Since agents share the nominal dynamics $f,g$ up to small parameter variations, their trajectories can be well approximated by common temporal features. We thus use the shared temporal basis
$\tilde\psi_\omega(t)\in\R^P$ from
\eqref{eq:network-output}, while agent-specific coefficients
$\tilde W_i\in\R^{P\times d}$, including the bias,
capture differences in parameters and initial states:
\begin{equation}
\label{eq:trajectory-representation}
  \tilde x_i(t)=\tilde W_i^\top\tilde\psi_\omega(t),
  \qquad
  \dot{\tilde x}_i(t)=\tilde W_i^\top\dot{\tilde\psi}_\omega(t).
\end{equation}
Thus $\tilde W=[\tilde W_1,\ldots,\tilde W_N]$ gives $\tilde x=\tilde W^\top\tilde\psi_\omega$. Basis values and derivatives are evaluated once for all agents; only the output layer grows with $N$. For the pairwise model \eqref{eq:agent-dynamics}, agent $i$'s residual is
\begin{equation}
\label{eq:forward-residual}
  r_i(t)=\tilde W_i^\top\dot{\tilde\psi}_\omega(t)
  -f(\tilde x_i(t);\alpha_i)
  -\sum_{j\in\mathcal N_i}g(\tilde x_i(t),\tilde x_j(t);\beta_{ij}).
\end{equation}
At fixed $\theta$, each residual depends on its own and neighbors' coefficients, so they are fitted jointly, coupled along the graph. With $r_{\mathrm{phys}}=\col(r_1,\ldots,r_N)$ and observations \eqref{eq:observations}, we minimize
\begin{equation}
\label{eq:inverse-loss}
\begin{aligned}
  \mathcal L_{\mathrm{inv}}(\omega,\tilde W,\theta)
  ={}&\frac{1}{N_{\mathrm d}}\sum_{m=1}^{N_{\mathrm d}}
    \|\mathcal C\tilde W^\top\tilde\psi_\omega(t_m)-y_m\|^2+\frac{\lambda_f}{N_f}\sum_{\ell=1}^{N_f}
    \|r_{\mathrm{phys}}(\tau_\ell;\omega,\tilde W,\theta)\|^2.
\end{aligned}
\end{equation}
The observation term covers measured components, and the physics term covers all agents; as both train the same basis, features learned from observed agents also serve unobserved ones. Each outer iteration $s$ updates in sequence three blocks that mirror the MAS structure: the shared basis $\omega$, the agent-specific coefficients $\tilde W$ coupled only through neighbors, and the time-invariant parameters $\theta$.

First, we update $\omega$ at fixed $(\tilde W^{(s)},\theta^{(s)})$. A gradient step takes the form
\begin{equation}
\label{eq:representation-update}
  \omega^{(s+1)}=\omega^{(s)}-\eta_\omega
  \nabla_\omega\mathcal L_{\mathrm{inv}}
  (\omega^{(s)},\tilde W^{(s)},\theta^{(s)}),
\end{equation}
with learning rate $\eta_\omega>0$. Through the physics loss, the basis adapts to the current $\theta^{(s)}$.

Second, we fix $(\omega^{(s+1)},\theta^{(s)})$ and fit $\tilde W$ using LM. Let $\gamma=\vecop(\tilde W)\in\R^{Pn}$, with $\vecop$ stacking columns. The residual vector $e(\gamma)$ stacks observation and physics residuals weighted by $N_{\mathrm d}^{-1/2}$ and $\sqrt{\lambda_f/N_f}$ with its explicit
form given in \eqref{eq:inverse-stacked-residual}. Then $\mathcal L_{\mathrm{inv}}=\|e(\gamma)\|^2$, where nonlinear dynamics generally make $e$ nonlinear in $\gamma$. Starting from $\gamma_0=\vecop(\tilde W^{(s)})$, each LM iteration $j$ computes
\begin{equation}
\label{eq:lm-update}
  (J_j^\top J_j+\mu_j I)\Delta\gamma_j=-J_j^\top e_j,
  \qquad e_j=e(\gamma_j),\quad J_j=\frac{\partial e}{\partial\gamma}(\gamma_j),
\end{equation}
where $\mu_j>0$. For an accepted step, $\gamma_{j+1}=\gamma_j+\Delta\gamma_j$.
The final coefficient is reshaped to $\tilde W^{(s+1)}$.

Finally, with $\omega^{(s+1)}$ and $\tilde W^{(s+1)}$
fixed, we update $\theta$ by approximately solving
the least-squares
\begin{equation}
\label{eq:parameter-subproblem}
  \min_\theta\frac{\lambda_f}{N_f}
  \sum_{\ell=1}^{N_f}
  \|r_{\mathrm{phys}}(\tau_\ell;
  \omega^{(s+1)},\tilde W^{(s+1)},\theta)\|^2,
\end{equation}
initialized at $\theta^{(s)}$ to obtain $\theta^{(s+1)}$; the observation term is independent of $\theta$.
Here, agents are decoupled, and each parameter enters only the residuals it acts on, making this subproblem sparse.

\paragraph{Matrix-free implementation.}
The matrix in \eqref{eq:lm-update} has dimension $PNd$, so assembling it is prohibitive for large $N$, yet the MAS structure makes it sparse: two agents are coupled in $J_j^\top J_j$ only if they enter a common residual, i.e., are at most two hops apart in the graph. The damped LM system is symmetric positive definite for $\mu_j>0$
(Appendix~\ref{app:structured-cost}). We therefore solve it approximately
by conjugate gradients (CG) method~\citep{hestenes1952methods}, starting from a
zero increment. CG only requires $-J_j^\top e_j$ and products $(J_j^\top J_j+\mu_j I)v=J_j^\top(J_jv)+\mu_jv$ for $v\in\R^{Pn}$, computed without assembling $J_j$ or $J_j^\top J_j$.  For readability, we drop the index $j$ below.

The product $Jv$ gives the first-order change in the residual along $v$. For each agent $i$, arrange its entries of $v$ column-wise into $V_i\in\R^{P\times d}$ as $\tilde W_i$. With the basis and physical parameters fixed, let $B_i,H_{ij}\in\R^{d\times d}$ be the derivatives of the full dynamics
$f(x_i;\alpha_i)+\sum_{j\in\mathcal N_i}g(x_i,x_j;\beta_{ij})$
with respect to $x_i$ and $x_j$, respectively, evaluated at the current reconstructed states. The chain rule maps coefficient perturbations to state and residual changes:
\begin{equation}
\label{eq:structured-jvp}
\begin{aligned}
  \delta x_i&=V_i^\top\tilde\psi_\omega, \quad
  \delta\dot x_i=V_i^\top\dot{\tilde\psi}_\omega,\\
  \delta r_i&=\delta\dot x_i-B_i\delta x_i
    -\sum_{j\in\mathcal N_i}H_{ij}\delta x_j.
\end{aligned}
\end{equation}
The physics and observation blocks of $Jv$ are $\sqrt{\lambda_f/N_f}\,\delta r_i(\tau_\ell)$ and $N_{\mathrm d}^{-1/2}\mathcal C\delta x(t_m)$, respectively. Computing $Jv$ thus mirrors the MAS itself: one message-passing step over the graph. The shared basis maps all $V_i$ to state perturbations in a single matrix product, and each agent combines its own and its neighbors' perturbations through $B_i,H_{ij}$, which can be batched as agents share $f$ and $g$.

Conversely, for $u$ structured as $e$, $J^\top u$ passes messages backward, returning each physics-residual block to its own and neighbors' coefficients, and each observation block via $\mathcal C^\top$ and the basis with weight $N_{\mathrm d}^{-1/2}$. The joint LM solve allows observation information to influence unobserved agents through the couplings induced by the physics residuals. Taking $u=Jv$ gives $J^\top Jv$, and $u=-e$ yields $-J^\top e$ in \eqref{eq:lm-update} (Appendix~\ref{app:structured-cost}). The computational savings come from evaluating $Jv$ and $J^\top u$ through the shared basis and the cached state-Jacobian blocks. At each collocation time, the physics part of each product applies one $B_i$ block per agent and one $H_{ij}$ block per modeled dependency. Thus, when each agent depends on only a bounded number of other agents on average, the work for these block multiplications grows linearly with $N$.

\begin{remark}
    For higher-order interactions, $\mathcal N_i$ includes every other agent entering agent $i$'s dynamics, and $B_i,H_{ij}$ include all interaction terms. The products \(Jv\) and \(J^\top u\) are evaluated using the same formulas with these expanded neighbor sets and derivative blocks.
\end{remark}

\subsection{Inference as a special case}
\label{sec:forward}

For inference, the physical parameters $\theta$ and initial state $x_0=x(0)$ are given. The same training framework applies by fixing $\theta$ and replacing the observation data with the prescribed initial state: $N_{\mathrm d}=1$, $t_1=0$, $\mathcal C=I_n$, and $y_1=x_0$. The objective \eqref{eq:inverse-loss} then reduces to
\begin{equation}
\label{eq:forward-loss}
  \mathcal L_{\mathrm{fwd}}(\omega,\tilde W)
  =\|\tilde W^\top\tilde\psi_\omega(0)-x_0\|^2
  +\frac{\lambda_f}{N_f}\sum_{\ell=1}^{N_f}
    \|r_{\mathrm{phys}}(\tau_\ell;\omega,\tilde W,\theta)\|^2.
\end{equation}
We alternate the basis and coefficient updates above using $\mathcal L_{\mathrm{fwd}}$, and omit the physical-parameter update. The residual vector now stacks the initial-state residual with unit weight and the same weighted physics residuals, as given in \eqref{eq:forward-stacked-residual}. The matrix-free physics products are unchanged, while the observation block of $Jv$ becomes the stacked $\delta x_i(0)$. Thus, inference uses the same shared representation and coupled LM solve.

\subsection{Computational complexity}
\label{sec:complexity}

To show that MAS-PINN reduces memory usage, we bound the cost of the structured coefficient solve after caching basis values, time derivatives, and local state-Jacobian blocks.
Let $\E=\{(j,i):j\in\mathcal N_i\}$, where
$(j,i)$ denotes that agent $i$'s dynamics depend on agent $j$'s state.
Thus $|\E|=\sum_{i=1}^N|\mathcal N_i|$ counts the off-diagonal
state-Jacobian blocks $H_{ij}$ at each collocation time.
For identification, $\E$ is defined by the candidate interactions
included in the model.

\begin{proposition}[Computational cost]
\label{prop:structured-cost}
With these cached quantities and sparse storage of $\mathcal C$, one application of $v\mapsto J^\top(Jv)+\mu v$ for identification costs
\begin{equation}
\label{eq:structured-cost}
  O\!\Big(N_f[NPd+(N+|\E|)d^2]
    +N_{\mathrm d}[NPd+\operatorname{nnz}(\mathcal C)]\Big),
\end{equation}
where $\operatorname{nnz}$ counts nonzero entries. For inference, the observation contribution is $O(NPd)$. For fixed $P,d,N_f,N_{\mathrm d}$, for sparse $\E$ ($|\E|=O(N)$) and, if
$\mathcal C$ selects distinct state components, each application
requires $O(N)$ work and the solve uses $O(N)$ storage, including caches.

\end{proposition}

The coefficient system has dimension $PNd$: dense normal-matrix storage and Cholesky factorization require $O((PNd)^2)$ memory and $O((PNd)^3)$ work. Under the proposition's assumptions, the structured solve instead uses $O((k_{\mathrm{CG}}+1)N)$ work and $O(N)$ storage for $k_{\mathrm{CG}}$ CG iterations, including formation of $-J^\top e$. This is a bound for a cached linear solve, not the full training procedure, and does not assume that $k_{\mathrm{CG}}$ is independent of $N$. The proof is in Appendix~\ref{app:structured-cost}.

\section{Numerical simulations}
\label{sec:experiments}

\paragraph{Objective.}
We conduct numerical experiments to evaluate the advantages
of MAS-PINN in modeling MAS.
First, we assess its ability to accurately infer the
trajectories of large-scale systems.
Second, we evaluate joint trajectory reconstruction and
parameter recovery in inverse problems involving partial
observations or unknown interaction topology.
Finally, we examine the computational efficiency of the
structured coefficient update, comparing its runtime and
memory requirements with dense LM and assessing its
scaling with system size.

\paragraph{Experiment setup.}
For inference, we compare MAS-PINN with vanilla PINN and
PirateNets~\citep{wang2024piratenets}.
For identification, the baselines are HyperPINN
\citep{mao2025hyperpinn} and an adaptive moving-horizon
estimator (MHE) \citep{schiller2026nonlinear}.
Methods that employ a standard multi-layer perceptron use a fully connected network with $\tanh$ activations and Adam~\citep{kingma2014adam} for gradient-based updates. For the physics-informed methods, collocation points are uniformly spaced and $\lambda_f=0.01$.
Each method is evaluated over five independent trials with different random seeds. We measure trajectory and parameter accuracy by the relative $L^2$ error
$\norm{\tilde{x}-x}_2/\norm{x}_2$ and $\norm{\tilde{\theta}-\theta}_2/\norm{\theta}_2$.
Errors for individual parameter blocks and unobserved
trajectories are reported where applicable.
Neural methods run on an NVIDIA RTX 4090 GPU, whereas
MHE runs on an Intel Core i9-13900K CPU.
Implementation details and solver settings for the experiments are provided in Appendix~\ref{app:settings}. The code for all benchmarks is available at \url{https://github.com/CyberAltrumi/MAS-PINN}.

\subsection{Inference for the Lorenz model on a ring}
\label{sec:lorenz96}
The Lorenz model \citep{lorenz1996predictability} describes
$N$ sites arranged on a ring, each with a scalar state.
The dynamics combine quadratic interactions between neighboring
states, linear damping, and a constant forcing $F$. The dynamics of site $i$ are given by
\begin{equation}
\label{eq:lorenz96}
  \dot x_i=(x_{i+1}-x_{i-2})x_{i-1}-x_i+F,\qquad i\in\V,
\end{equation}
with cyclic indices. The local dynamics are $f(x_i)=-x_i+F$, while the advection term couples site
$i$ to $\mathcal N_i=\{i-2,i-1,i+1\}$ through a product of neighbor states. We consider the inference problem with $N\in\{50,100\}$ sites and
$F=12$, given the initial state of every site on the attractor, and approximate the trajectory over $[0,1]$. The initial conditions and algorithmic settings are listed in Table~\ref{tab:settings-lorenz}.

Table~\ref{tab:lorenz-results} reports the relative $L^2$ errors
of the trajectory over $[0,1]$ and the final state $x(1)$.
For both network sizes, MAS-PINN achieves mean trajectory errors at least two orders of magnitude lower
than those of baselines with fewer training epochs.
Both trajectory and final-state errors for MAS-PINN remain comparable
as $N$ increases from $50$ to $100$.
Figure~\ref{fig:lorenz-errors} presents the corresponding
spatiotemporal error fields for $N=100$.
These results demonstrate accurate trajectory inference of MAS-PINN at large network sizes.

\begin{table}[hb]
\centering
\caption{Inference results for the Lorenz experiment. Entries are
the mean $\pm$ standard deviation over five independent trials of the relative
$L^2$ errors of the trajectory over $[0,1]$ and of the final state
$x(1)$. Best mean values are in bold.}
\label{tab:lorenz-results}
\small
\setlength{\tabcolsep}{4pt}
\begin{tabular}{@{}lcccc@{}}
\toprule
& \multicolumn{2}{c}{$N=50$} & \multicolumn{2}{c}{$N=100$}\\
\cmidrule(lr){2-3}\cmidrule(l){4-5}
Method & Trajectory & Final state & Trajectory & Final state\\
\midrule
Vanilla PINN & $(6.0 \pm 1.3)\times 10^{-1}$ & $(9.6 \pm 1.0)\times 10^{-1}$ & $(5.6 \pm 0.9)\times 10^{-1}$ & $(9.1 \pm 1.7)\times 10^{-1}$\\
PirateNets & $(1.2 \pm 0.8)\times 10^{-1}$ & $(3.3 \pm 2.4)\times 10^{-1}$ & $(1.6 \pm 0.7)\times 10^{-1}$ & $(3.2 \pm 2.2)\times 10^{-1}$\\
MAS-PINN & $\mathbf{(1.9 \pm 1.1)\times 10^{-4}}$ & $\mathbf{(9.1 \pm 6.0)\times 10^{-4}}$ & $\mathbf{(2.2 \pm 0.9)\times 10^{-4}}$ & $\mathbf{(7.9 \pm 5.3)\times 10^{-4}}$\\
\bottomrule
\end{tabular}
\end{table}

\subsection{Partially observed network Lotka--Volterra dynamics}
\label{sec:network-lv}
The networked Lotka--Volterra model describes an ecological
network of $N$ sites, with local interactions among three
species at each site and dispersal between sites. We assume that the network topology and the diagonal diffusion
matrix $D$ are known. The dynamics at site $i$ are given by
\begin{equation}
\label{eq:network-lv}
  \dot{x}_i
  = x_i \odot  A x_i
    + \sum_{j\in\mathcal N_i} w_{ij}D(x_j-x_i),\qquad i\in\V.
\end{equation}
where $x_i\in\mathbb R^3$ collects the species densities at
site $i$, and $\odot$ denotes elementwise multiplication.
The matrix $A\in\mathbb R^{3\times3}$ characterizes local
interactions among species, while $w_{ij}=w_{ji}$ determines the dispersal strength along the undirected edge $\{i,j\}$. The interaction matrix $A$ is shared across sites, giving
$\theta=\col(\vecop(A),w)$, where $w$ collects one weight
per undirected edge in a fixed order.

\begin{figure}[t]
  \centering
  \includegraphics[width=\linewidth,trim=0 10 0 6,clip]{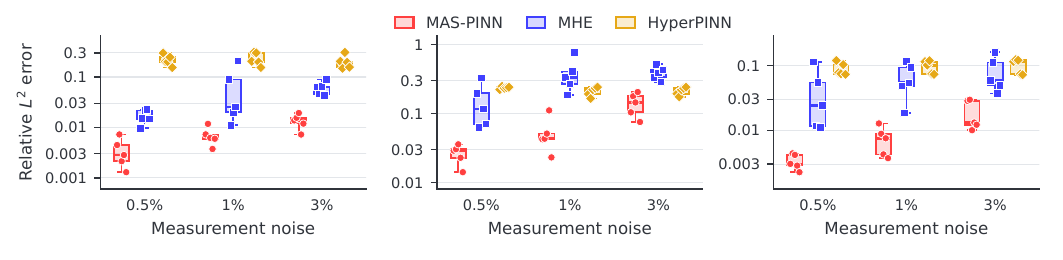}
  \caption{Relative $L^2$ error versus measurement noise for all methods. Left: local
interaction matrix $A$; middle: dispersal weights $w$; right:
all site trajectories. Boxes show
interquartile ranges and medians; whiskers reach the extreme values
within $1.5$ interquartile ranges of the box. Points show all five trials.}
  \label{fig:lv-errors}
\end{figure}

We consider $N=8$ sites and identify $A$ and $w$ from
noisy observations of all three species at half of the sites.
The ground-truth parameters, observations, and
algorithmic settings are listed in Table~\ref{tab:settings-lv}.
Figure~\ref{fig:lv-errors} reports the relative $L^2$ errors
in estimating $A$ and $w$ and reconstructing trajectories
at all sites.
Across all three noise levels, MAS-PINN achieves lower
errors than baselines for all three quantities.
Representative trajectory reconstructions at $0.5\%$ noise
are shown in Figure~\ref{fig:lv-trajectories}.
These results support the effectiveness of MAS-PINN in joint parameter identification and unobserved-state reconstruction from partial observations.

\begin{figure}[t]
\centering
\includegraphics[width=0.667\linewidth,trim=0 5 0 7,clip]{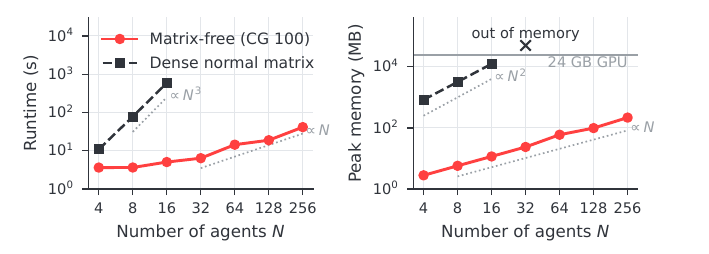}
\caption{Coefficient update scaling with $N$ ($N_f=1600$, $P=129$, 100 CG iterations).
Left: LM time over 100 outer iterations. Right: peak LM-stage GPU memory.
Dotted lines show reference slopes. Dense LM was not run for $N\ge32$,
as its estimated memory ($\times$ at $N=32$) exceeds 24\,GB.}
\label{fig:lv-cost}
\end{figure}

\paragraph{Computational cost.}
We compare the matrix-free coefficient update with a dense
LM implementation that explicitly forms the Jacobian and
normal matrix and uses Cholesky factorization.
At the reference setting, the matrix-free implementation
substantially reduces runtime and peak GPU memory
(Table~\ref{tab:cost-lv}).
With $N_f$, $P$, and the CG iteration budget fixed,
Figure~\ref{fig:lv-cost} shows approximately linear growth
in coefficient-update time and memory at larger network sizes, consistent with the analysis
in Proposition~\ref{prop:structured-cost}.
These results support the computational efficiency and
scalability of the matrix-free coefficient update.
Additional scaling results are in
Table~\ref{tab:scaling-lv} in the appendix.

\begin{table}[t]
\centering
\caption{Computational cost at the reference setting
($N=8$, $N_f=1600$, $P=129$).
Times are measured over 100 outer iterations, peak GPU memory covers the full iteration.
LM denotes the coefficient update, and Total includes the outer iteration.}
\label{tab:cost-lv}
\small
\begin{tabular}{@{}lrrr@{}}
\toprule
Coefficient solver
& LM (s)
& Total (s)
& Peak memory (MB)\\
\midrule
Matrix-free, 30 CG iterations
& 1.20 & 1.78 & 32\\
Matrix-free, 60 CG iterations
& 2.25 & 2.82 & 32\\
Matrix-free, 100 CG iterations
& 3.64 & 4.26 & 32\\
Dense LM, Cholesky
& 77.77 & 78.44 & 3194\\
\bottomrule
\end{tabular}
\end{table}

\subsection{Rössler oscillators on undirected hypergraphs}
\label{sec:rossler-hypergraph}
The coupled Rössler model describes nonlinear oscillators interacting
through pairwise and three-body couplings on an undirected hypergraph. The dynamics of oscillator $i$
are given by
\begin{equation}
\label{eq:rossler-hypergraph}
\begin{aligned}
  \dot x_{i,1} &= -x_{i,2}-x_{i,3}
   +\sum_{j\in\mathcal N_i} w_{ij}(x_{j,1}-x_{i,1})
    +\sum_{\{j,k\}\in\mathcal T_i} v_{ijk}
     \bigl(x_{j,1}^2x_{k,1}+x_{j,1}x_{k,1}^2-2x_{i,1}^3\bigr),\\
  \dot x_{i,2} &= x_{i,1}+ax_{i,2},\qquad
  \dot x_{i,3} = b+x_{i,3}(x_{i,1}-c),\qquad i\in\V.
\end{aligned}
\end{equation}
where $x_i=\col(x_{i,1},x_{i,2},x_{i,3})$, and \(\mathcal N_i\) and \(\mathcal T_i\) contain candidate neighbors and unordered pairs of distinct nodes other than \(i\), respectively. The local parameters \(a,b,c\) are known, whereas the topology and weights are unknown; each undirected hyperedge has one weight, zero if absent.

Following \citet{mao2025hyperpinn}, we consider $N=8$
oscillators, with all $28$ pairs and $56$ triples
as candidate interactions.
All states are
observed with measurement noise. We also evaluate topology
identification using the area under the receiver operating
characteristic curve (AUC).
We pool all candidate pairwise and three-body interactions,
with identified weights as scores and
ground-truth presence as binary labels. The ground truth, training data, and algorithmic settings
are in Table~\ref{tab:settings-rossler}.

Table~\ref{tab:rossler-results} summarizes the results.
MAS-PINN and MHE achieve an AUC of $1$ at all noise
levels, whereas HyperPINN has lower AUC and large
three-body weight errors.
MAS-PINN also achieves the lowest relative $L^2$
errors in weights and trajectories at every noise level, though they grow with noise.
Figures~\ref{fig:rossler-errors}--\ref{fig:rossler-trajectories} (appendix) show error distributions, ROC curves, and $3\%$-noise reconstructions.

\begin{table}[t]
\centering
\caption{Rössler hypergraph identification results.
Entries report mean $\pm$ standard deviation over five trials
for relative $L^2$ errors in $w$, $v$, and trajectories,
and AUC over all candidate hyperedges.
AUC of $1.00$ means unity in every trial.
Best means per noise level are bold.}
\label{tab:rossler-results}
\small
\setlength{\tabcolsep}{4pt}
\begin{tabular}{@{}llrrrr@{}}
\toprule
Noise & Method & $w$ & $v$ & Trajectory & AUC\\
\midrule
$0.5\%$ & MAS-PINN & $\mathbf{(4.5 \pm 2.3)\times 10^{-3}}$ & $\mathbf{(9.7 \pm 7.0)\times 10^{-3}}$ & $\mathbf{(2.48 \pm 0.04)\times 10^{-3}}$ & $\mathbf{1.00}$\\
        & HyperPINN   & $(4.7 \pm 2.6)\times 10^{-1}$ & $(9.9 \pm 0.1)\times 10^{-1}$ & $(9.3 \pm 1.7)\times 10^{-2}$ & $0.96 \pm 0.02$\\
        & MHE         & $(3.2 \pm 1.6)\times 10^{-2}$ & $(2.8 \pm 1.6)\times 10^{-2}$ & $(6.8 \pm 3.8)\times 10^{-3}$ & $\mathbf{1.00}$\\
\midrule
$1\%$   & MAS-PINN & $\mathbf{(8.5 \pm 4.6)\times 10^{-3}}$ & $\mathbf{(1.8 \pm 1.2)\times 10^{-2}}$ & $\mathbf{(4.9 \pm 0.1)\times 10^{-3}}$ & $\mathbf{1.00}$\\
        & HyperPINN   & $(4.7 \pm 2.6)\times 10^{-1}$ & $(9.9 \pm 0.1)\times 10^{-1}$ & $(9.4 \pm 1.6)\times 10^{-2}$ & $0.96 \pm 0.02$\\
        & MHE         & $(5.4 \pm 1.7)\times 10^{-2}$ & $(4.8 \pm 2.5)\times 10^{-2}$ & $(1.2 \pm 0.6)\times 10^{-2}$ & $\mathbf{1.00}$\\
\midrule
$3\%$   & MAS-PINN & $\mathbf{(2.7 \pm 1.5)\times 10^{-2}}$ & $\mathbf{(6.6 \pm 3.2)\times 10^{-2}}$ & $\mathbf{(1.5 \pm 0.1)\times 10^{-2}}$ & $\mathbf{1.00}$\\
        & HyperPINN   & $(4.7 \pm 2.6)\times 10^{-1}$ & $(9.9 \pm 0.1)\times 10^{-1}$ & $(9.3 \pm 1.8)\times 10^{-2}$ & $0.96 \pm 0.02$\\
        & MHE         & $(1.1 \pm 0.2)\times 10^{-1}$ & $(1.0 \pm 0.5)\times 10^{-1}$ & $(2.6 \pm 1.1)\times 10^{-2}$ & $\mathbf{1.00}$\\

\bottomrule
\end{tabular}
\end{table}

\section{Conclusion}
\label{sec:conclusion}
We introduced MAS-PINN, a structure-aware
physics-informed framework for inference and
identification in MAS.
Combining a shared temporal representation,
alternating optimization, and matrix-free coefficient
updates, it exploits local state dependencies
to reduce the cost of coupled trajectory fitting.
Experiments show accurate
inference in large MAS and joint state
and parameter recovery under partial observations
or unknown topology, with lower runtime and memory than dense LM.
Future work will extend MAS-PINN beyond the fixed-parameter, offline setting, to time-varying
interaction structures and online state and parameter
estimation from sequential observations.

\section*{Acknowledgments}
This work was partially supported by the Wallenberg AI, Autonomous Systems
and Software Program (WASP), funded by the Knut and Alice Wallenberg
Foundation. This work was also supported by the Swedish Research Council
under Distinguished Professor Grant 2017-01078. During the preparation of this manuscript, the authors used ChatGPT (OpenAI)
from July to October 2026 to improve the clarity of selected passages,
identify potential weaknesses in the draft, and help condense the text.
The authors reviewed the suggestions, edited and approved the final
manuscript, and take full responsibility for its content.

\bibliographystyle{unsrtnat}
\bibliography{references}

\appendix
\section{Structured products and computational complexity}
\label{app:structured-cost}

Throughout this appendix, $\omega$, $\theta$, and the LM
linearization point are fixed.
The operation and storage counts concern the linear solve
after caching the basis values, basis derivatives, and
state-Jacobian blocks.

\paragraph{Identification residual and LM step.}
At outer iteration $s$, coefficient fitting fixes
$\omega=\omega^{(s+1)}$ and $\theta=\theta^{(s)}$.
With $\gamma=\vecop(\tilde W)$, the identification residual is
\begin{equation}
\label{eq:inverse-stacked-residual}
  e(\gamma)=
  \begin{bmatrix}
    N_{\mathrm d}^{-1/2}
    (\mathcal C\tilde W^\top\tilde\psi_{\omega^{(s+1)}}(t_1)-y_1)\\
    \vdots\\
    N_{\mathrm d}^{-1/2}
    (\mathcal C\tilde W^\top\tilde\psi_{\omega^{(s+1)}}(t_{N_{\mathrm d}})-y_{N_{\mathrm d}})\\
    \sqrt{\lambda_f/N_f}\,
    r_{\mathrm{phys}}(\tau_1;\omega^{(s+1)},\tilde W,\theta^{(s)})\\
    \vdots\\
    \sqrt{\lambda_f/N_f}\,
    r_{\mathrm{phys}}(\tau_{N_f};\omega^{(s+1)},\tilde W,\theta^{(s)})
  \end{bmatrix}.
\end{equation}
With $\gamma_j$ the current LM iterate, define
$e_j=e(\gamma_j)$ and
$J_j=\partial e(\gamma_j)/\partial\gamma$.
The damped linearized least-squares problem is
\begin{equation}
\label{eq:lm-subproblem}
  \Delta \gamma_j
  =
  \arg\min_{\Delta}
  \left\{
    \frac12\|e_j+J_j\Delta\|^2
    +\frac{\mu_j}{2}\|\Delta\|^2
  \right\}.
\end{equation}
Setting its gradient to zero gives
$J_j^\top(e_j+J_j\Delta\gamma_j)+\mu_j\Delta\gamma_j=0$,
which yields \eqref{eq:lm-update}.
For $\mu_j>0$ and every nonzero coefficient-space vector $v$,
\[
  v^\top(J_j^\top J_j+\mu_j I)v
  =\|J_jv\|^2+\mu_j\|v\|^2>0.
\]
Hence the damped normal matrix is symmetric positive definite,
as required by CG. Its action on $v$ is obtained by composing
$J_jv$ with $J_j^\top$ and adding $\mu_jv$; the structured
products below evaluate these operations without assembling
the coefficient Jacobian.

\paragraph{Inference specialization.}
Fixing the prescribed $\theta$ and setting $N_{\mathrm d}=1$,
$t_1=0$, $\mathcal C=I_n$, and $y_1=x_0$ replaces the
observation blocks by the initial-state residual. Under this specialization, the residual becomes
\begin{equation}
\label{eq:forward-stacked-residual}
  e(\gamma)=
  \begin{bmatrix}
    \tilde{W}^\top\tilde{\psi}_{\omega^{(s+1)}}(0)-x_0\\
    \sqrt{\lambda_f/N_f}\,
    r_{\mathrm{phys}}(\tau_1;\omega^{(s+1)},\tilde{W},\theta)\\
    \vdots\\
    \sqrt{\lambda_f/N_f}\,
    r_{\mathrm{phys}}(\tau_{N_f};\omega^{(s+1)},\tilde{W},\theta)
  \end{bmatrix}.
\end{equation}
The same LM derivation applies to this residual.

\paragraph{Structured Jacobian products.}
For the pairwise model \eqref{eq:agent-dynamics}, define
$\Phi_i(x;\theta)=f(x_i;\alpha_i)
+\sum_{j\in\mathcal N_i}g(x_i,x_j;\beta_{ij})$.
For higher-order models, $\Phi_i$ includes all interaction terms. Then we have
\[
  r_i(t)=\dot{\tilde x}_i(t)
  -\Phi_i(\tilde x(t);\theta).
\]
Define the structural state dependencies by
\[
  \mathcal N_i
  =\{j\ne i:\Phi_i\text{ structurally depends on }x_j\},
  \qquad
  \E=\{(j,i):j\in\mathcal N_i\},
\]
and let
\[
  B_i=\frac{\partial\Phi_i}{\partial x_i},
  \qquad
  H_{ij}=\frac{\partial\Phi_i}{\partial x_j},
  \quad j\in\mathcal N_i.
\]
These derivatives are evaluated at the current reconstructed
states and physical parameters.

Column-wise vectorization gives
\[
  \tilde x_i(t)
  =(I_d\otimes\tilde\psi_\omega(t)^\top)
    \vecop(\tilde W_i),
\]
with the analogous expression for $\dot{\tilde x}_i$
using $\dot{\tilde\psi}_\omega$.
Here $\otimes$ denotes the Kronecker product.
The elementary derivatives are
\begin{equation}
\label{eq:elementary-derivatives}
\begin{aligned}
  \frac{\partial\tilde x_i}{\partial\vecop(\tilde W_i)}
  &=I_d\otimes\tilde\psi_\omega^\top,
  &
  \frac{\partial\dot{\tilde x}_i}{\partial\vecop(\tilde W_i)}
  &=I_d\otimes\dot{\tilde\psi}_\omega^\top,\\
  \frac{\partial r_i}{\partial\tilde x_i}
  &=-B_i,
  &
  \frac{\partial r_i}{\partial\tilde x_j}
  &=-H_{ij},
  \quad j\in\mathcal N_i.
\end{aligned}
\end{equation}
Applying the chain rule gives the unweighted
physics-Jacobian blocks
\begin{equation}
\label{eq:jacobian-blocks}
  \frac{\partial r_i}{\partial\vecop(\tilde W_j)}
  =
  \begin{cases}
    I_d\otimes\dot{\tilde\psi}_\omega^\top
      -B_i\otimes\tilde\psi_\omega^\top,
      &j=i,\\[1mm]
    -H_{ij}\otimes\tilde\psi_\omega^\top,
      &j\in\mathcal N_i,\\[1mm]
    0, &\text{otherwise}.
  \end{cases}
\end{equation}
Each block belongs to $\R^{d\times Pd}$.
Multiplying \eqref{eq:jacobian-blocks} by
$\vecop(V_j)$ and summing over $j$ yields
\eqref{eq:structured-jvp}.

For the transpose product, let $u_i\in\R^d$ be the physics
block of a residual-space vector $u$ at one collocation time.
The contribution to agent $i$'s coefficient matrix is
\begin{equation}
\label{eq:structured-vjp}
\begin{aligned}
  Z_i=\sqrt{\lambda_f/N_f}\,\Big[
  \dot{\tilde\psi}_\omega u_i^\top-\tilde\psi_\omega
  \Big(B_i^\top u_i+
  \sum_{h:\,i\in\mathcal N_h}H_{hi}^\top u_h\Big)^\top
  \Big].
\end{aligned}
\end{equation}
Summing $\vecop(Z_i)$ over collocation times gives the physics
part of the agent-$i$ block of $J^\top u$.
To verify the transpose identity
$\langle u,Jv\rangle=\langle J^\top u,v\rangle$,
group all terms involving the same $V_i$:
\[
\begin{aligned}
  \sqrt{\lambda_f/N_f}\sum_i u_i^\top\delta r_i&=
  \sqrt{\lambda_f/N_f}\sum_i
  \Big[
    u_i^\top V_i^\top\dot{\tilde\psi}_\omega
    -\Big(
      B_i^\top u_i+
      \sum_{h:\,i\in\mathcal N_h}H_{hi}^\top u_h
     \Big)^\top V_i^\top\tilde\psi_\omega
  \Big]\\
  &=\sum_i\langle V_i,Z_i\rangle_F,
\end{aligned}
\]
where
$\langle U,V\rangle_F=\operatorname{tr}(U^\top V)$
is the Frobenius inner product and $Z_i$ is given by
\eqref{eq:structured-vjp}.
The sum over $h$ collects the contributions from every
residual that depends on the state of agent $i$.

At observation time $t_m$, the weighted Jacobian is
\[
  N_{\mathrm{d}}^{-1/2}\,
  \mathcal C
  (I_n\otimes\tilde\psi_\omega(t_m)^\top).
\]

Let $u_m\in\R^{n_s}$ be the observation block of $u$
at time $t_m$. Applying the transpose of the observation
Jacobian contributes
\[
  N_{\mathrm d}^{-1/2}\,
  \tilde\psi_\omega(t_m)
  (\mathcal C^\top u_m)_i^\top
\]
to agent $i$'s coefficient matrix, where
$(\mathcal C^\top u_m)_i\in\R^d$ denotes the block
corresponding to agent $i$.
Vectorizing these matrix contributions and summing over
$m$ gives the observation part of the agent-$i$ block
of $J^\top u$.
For inference, the initial-state Jacobian is
$I_n\otimes\tilde\psi_\omega(0)^\top$.
Combining the data and physics contributions gives the
full weighted products $Jv$ and $J^\top u$.

\begin{proof}[Proof of Proposition~\ref{prop:structured-cost}]
Fix the basis parameters, physical parameters, and LM
linearization point. Assume that the basis values, time
derivatives, and state-Jacobian blocks are cached, and
that $\mathcal C$ is stored sparsely.

At one collocation time, forming all $\delta x_i$ and
$\delta\dot x_i$ costs $O(NPd)$ operations. Applying the
local blocks $B_i$ and coupling blocks $H_{ij}$ costs
$O((N+|\E|)d^2)$. The transpose product has the same order
of cost: the block multiplications cost
$O((N+|\E|)d^2)$, and the outer products in
\eqref{eq:structured-vjp} cost $O(NPd)$.

At each observation time, forming state perturbations
or coefficient contributions costs at most $O(NPd)$.
Multiplication by $\mathcal C$ or $\mathcal C^\top$
costs $O(\operatorname{nnz}(\mathcal C))$.
Summing over collocation and observation times, and
including the $O(PNd)$ damping operation, gives
\eqref{eq:structured-cost}. For inference, the observation
term is replaced by an initial-state contribution
of cost $O(NPd)$.

For storage, the state-Jacobian cache requires
$O(N_f(N+|\E|)d^2)$ entries, while the shared basis
values and derivatives require
$O((N_f+N_{\mathrm d})P)$ entries.
Including residual and observation vectors, the sparse
observation matrix, and a fixed number of coefficient-sized
CG work vectors gives
\begin{equation}
\label{eq:structured-storage}
\begin{aligned}
O\!\Bigl(
  &N_f(N+|\E|)d^2+(N_f+N_{\mathrm d})P+PNd+N_{\mathrm d}n_s
  +\operatorname{nnz}(\mathcal C)
\Bigr).
\end{aligned}
\end{equation}
Here the physics-residual storage $O(N_fNd)$ is absorbed
by the first term, and observation products are evaluated
one time point at a time. For inference, one initial-state
vector and the basis at $t=0$ replace the observation
data and basis values.

For observations selecting distinct state coordinates,
$n_s\le Nd$ and $\operatorname{nnz}(\mathcal C)=n_s$.
Together with $|\E|=O(N)$, these relations give $O(N)$
work per normal-operator application and $O(N)$ storage
for fixed $P,d,N_f,N_{\mathrm d}$.

Finally, constructing $-J^\top e$ requires one transpose
product. Each CG iteration requires one evaluation of
$J^\top(Jv)+\mu v$ and $O(PNd)$ vector operations.
Thus, $k_{\mathrm{CG}}$ iterations, including
right-hand-side construction, require
$O((k_{\mathrm{CG}}+1)N)$ work under the stated scaling
assumptions. Storage remains $O(N)$ because the work
vectors are reused across iterations.
\end{proof}

The physics Jacobian has $N_fNd$ rows and $PNd$ columns.
Dense storage therefore requires $O(N_fPN^2d^2)$ entries.
Storing every structurally present block as a dense
$d\times Pd$ matrix gives the graph-sparse bound
$O(N_f(N+|\E|)Pd^2)$.
The structured products retain the $d\times d$
state-Jacobian blocks and apply the shared basis separately,
avoiding their expansion into coefficient-space blocks.

The full normal matrix has size $PNd\times PNd$.
Dense storage requires $O((PNd)^2)$ entries, and dense
Cholesky factorization costs $O((PNd)^3)$ operations.
For fixed $P$ and $d$, these are $O(N^2)$ and $O(N^3)$,
respectively.

\section{Experimental settings}
\label{app:settings}
In this section, we provide case-specific experimental settings
and implementation details for MAS-PINN and all baseline methods. Table~\ref{tab:settings-updates} lists the shared optimization settings and Tables 5–7 the case-specific settings.

\begin{table}[htbp]
\centering
\caption{Shared optimization settings for MAS-PINN across all experiments.}
\label{tab:settings-updates}
\small
\renewcommand{\arraystretch}{1.2}
\begin{tabular}{@{}p{\dimexpr0.40\linewidth-\tabcolsep\relax}p{\dimexpr0.60\linewidth-\tabcolsep\relax}@{}}
\toprule
Setting & Value\\
\midrule
Updates per outer iteration
& 1 Adam step on $\omega$, 1 LM step on $\tilde W$,
and 1 parameter solve for identification only\\\midrule
Initial damping multiplier & $c_0=1\times10^{-3}$\\\midrule
LM damping & $\mu_j=c_j\,\operatorname{tr}(J_j^\top J_j)/(PNd)$\\\midrule
Accepted-step multiplier & $c_{j+1}=c_j/3$\\\midrule
Rejected-step multiplier & $c_{j+1}=2c_j$\\\midrule
Acceptance threshold & Actual decrease $\ge10^{-3}\times$ predicted decrease\\\midrule
Maximum CG iterations & 100\\\midrule
Relative CG residual tolerance & $1\times10^{-5}$\\
\bottomrule
\end{tabular}
\end{table}

\begin{table}[htbp]
\centering
\caption{Settings for the Lorenz inference experiment.
Following \citet{wang2024piratenets}, PirateNets use a random Fourier
embedding of time ($128$ frequencies, scale $2$) and are trained with
random weight factorization, physics-informed initialization, causal
weighting, gradient-norm loss balancing, a $5000$-step learning-rate
warm-up, and gradient clipping.
The reference trajectory is computed with DOP853 at relative and absolute tolerances $10^{-12}$.}

\label{tab:settings-lorenz}
\small
\renewcommand{\arraystretch}{1.2}
\begin{tabular}{@{}>{\raggedright\arraybackslash}p{\dimexpr0.24\linewidth-\tabcolsep\relax}>{\centering\arraybackslash}p{\dimexpr0.25\linewidth-2\tabcolsep\relax}>{\centering\arraybackslash}p{\dimexpr0.25\linewidth-2\tabcolsep\relax}>{\centering\arraybackslash}p{\dimexpr0.26\linewidth-\tabcolsep\relax}@{}}
\toprule
Setting & MAS-PINN & Vanilla PINN & PirateNets\\
\midrule
Time interval & \multicolumn{3}{p{\dimexpr0.76\linewidth-\tabcolsep\relax}@{}}{\centering $[0,1]$}\\\midrule
Sites & \multicolumn{3}{p{\dimexpr0.76\linewidth-\tabcolsep\relax}@{}}{\centering $N\in\{50,100\}$}\\\midrule
State dimension, forcing & \multicolumn{3}{p{\dimexpr0.76\linewidth-\tabcolsep\relax}@{}}{\centering $d=1$, $F=12$}\\\midrule
Initial state & \multicolumn{3}{p{\dimexpr0.76\linewidth-\tabcolsep\relax}@{}}{\centering Spin-up of $20$ time units from $x_i=F+0.01\,\xi_i$, $\xi_i\sim\mathcal N(0,1)$}\\\midrule
Collocation points & \multicolumn{3}{p{\dimexpr0.76\linewidth-\tabcolsep\relax}@{}}{\centering $N_f=1000$}\\\midrule
Network & MLP & MLP & 	PirateNet \\\midrule
Hidden depth & 3 & 5 & 5\\\midrule
Hidden width & 64 & 64 & 64\\\midrule
$D_f$ & 64 & --- & ---\\\midrule
$\sigma$ & $1$ & --- & ---\\\midrule
Training iterations & $1\times10^4$ & $2\times10^5$ & $2\times10^5$\\\midrule
Initial learning rate & \multicolumn{3}{p{\dimexpr0.76\linewidth-\tabcolsep\relax}@{}}{\centering $1\times10^{-3}$}\\\midrule
Decay rate & $0.97$ & $0.97$ & $0.9$\\\midrule
Decay step & $100$ epochs & $2000$ epochs & $5000$ epochs\\
\bottomrule
\end{tabular}
\end{table}

\begin{table}[htbp]
\centering
\caption{Settings for the partially observed network Lotka--Volterra experiment.
HyperPINN is trained on the data loss alone for $7000$ iterations,
after which the physics weight increases linearly to $1$ over
$7000$ iterations, with gradient clipping.
MHE solves a bounded nonlinear least-squares problem over a sliding
window of $81$ observations, with one RK4 step per sampling interval,
arrival-cost weight $1$, and parameter-prior weight $100$.}
\label{tab:settings-lv}
\small
\renewcommand{\arraystretch}{1.2}

\begin{tabular}{@{}p{\dimexpr0.34\linewidth-2\tabcolsep\relax}>{\centering\arraybackslash}p{\dimexpr0.33\linewidth-\tabcolsep\relax}>{\centering\arraybackslash}p{\dimexpr0.33\linewidth-\tabcolsep\relax}@{}}
\toprule
Setting & MAS-PINN & HyperPINN\\
\midrule
Time interval & \multicolumn{2}{p{0.66\linewidth}@{}}{\centering $[0,3]$}\\\midrule
Sites, species, diffusion & \multicolumn{2}{p{0.66\linewidth}@{}}{\centering $N=8$, $d=3$, $D=0.5\,I_3$}\\\midrule
Interaction matrix $A$ & \multicolumn{2}{p{0.66\linewidth}@{}}{\centering $[0,1,-1;\,-1,0,1;\,1,-1,0]$}\\\midrule
Graph & \multicolumn{2}{p{0.66\linewidth}@{}}{\centering Ring on the $8$ sites plus the edges $\{1,6\},\{3,5\},\{4,8\},\{5,7\}$}\\\midrule
Dispersal weights $w$ & \multicolumn{2}{p{0.66\linewidth}@{}}{\centering $(1.07,1.17,0.64,0.73,0.69,0.91,0.67,0.81,1.19,1.32,0.78,1.34)$}\\\midrule
Observed sites & \multicolumn{2}{p{0.66\linewidth}@{}}{\centering $\V_{\mathrm{o}}=\{1,3,5,7\}$}\\\midrule
Observation points & \multicolumn{2}{p{0.66\linewidth}@{}}{\centering $N_{\mathrm{d}}=160$}\\\midrule
Collocation points & \multicolumn{2}{p{0.66\linewidth}@{}}{\centering $N_f=1600$}\\\midrule
Measurement noise & \multicolumn{2}{p{0.66\linewidth}@{}}{\centering $\mathcal N(0,\eta^2)$,
$\eta\in\{0.005,0.01,0.03\}$}\\\midrule
Initial states & \multicolumn{2}{p{0.66\linewidth}@{}}{\centering $x_{i,k}(0)\sim\mathcal U(0.5,1.2)$,
$i\in\V$, $k=1,2,3$}\\\midrule
Hidden depth & 3 & 3\\\midrule
Hidden width & 64 & 128\\\midrule
$D_f$ & 64 & ---\\\midrule
$\sigma$ & $3$ & ---\\\midrule
Training iterations & $1\times10^4$ & $4\times10^4$\\\midrule
Initial learning rate & $1\times10^{-3}$ & $1\times10^{-3}$\\\midrule
Decay rate & $0.97$ & $0.95$\\\midrule
Decay step & $100$ epochs & $1000$ epochs\\
\bottomrule
\end{tabular}
\end{table}

\begin{table}[htbp]
\centering
\caption{Settings for the Rössler hypergraph experiment.
HyperPINN follows \citet{mao2025hyperpinn}: a residual network with
layer normalization, GELU activations, and dropout, sigmoid-gated
hyperedge weights with $L_1$/$L_0$ sparsity penalties, a three-stage
loss schedule, and AdamW with gradient clipping.
MHE uses the settings of Table~\ref{tab:settings-lv} with six RK4
steps per sampling interval.}
\label{tab:settings-rossler}
\small
\renewcommand{\arraystretch}{1.2}
\begin{tabular}{@{}p{\dimexpr0.34\linewidth-2\tabcolsep\relax}>{\centering\arraybackslash}p{\dimexpr0.33\linewidth-\tabcolsep\relax}>{\centering\arraybackslash}p{\dimexpr0.33\linewidth-\tabcolsep\relax}@{}}
\toprule
Setting & MAS-PINN & HyperPINN\\
\midrule
Time interval & \multicolumn{2}{p{0.66\linewidth}@{}}{\centering $[0,20]$}\\\midrule
Oscillators, local parameters & \multicolumn{2}{p{0.66\linewidth}@{}}{\centering $N=8$, $a=b=0.2$, $c=5.7$}\\\midrule
Pairwise interactions & \multicolumn{2}{p{0.66\linewidth}@{}}{\centering $\{1,2\},\{2,3\},\{3,4\},\{5,6\},\{6,7\},\{7,8\}$, strength $0.4$}\\\midrule
Three-body interactions & \multicolumn{2}{p{0.66\linewidth}@{}}{\centering $\{1,2,3\},\{2,4,5\},\{5,6,7\},\{6,7,8\}$, strength $0.3$}\\\midrule
Candidate hyperedges & \multicolumn{2}{p{0.66\linewidth}@{}}{\centering All $28$ pairs and $56$ triples}\\\midrule
Observations & \multicolumn{2}{p{0.66\linewidth}@{}}{\centering  $N_{\mathrm{d}}=151$}\\\midrule
Collocation points & \multicolumn{2}{p{0.66\linewidth}@{}}{\centering $N_f=151$ with same sampling time as observations}\\\midrule
Measurement noise & \multicolumn{2}{p{0.66\linewidth}@{}}{\centering $\mathcal N(0,\eta^2)$,
$\eta\in\{0.005,0.01,0.03\}$}\\\midrule
Initial states & \multicolumn{2}{p{0.66\linewidth}@{}}{\centering
\(x_{i,k}(0)\sim\mathcal U(-1,1)\),
$i\in\V$, $k=1,2,3$}\\\midrule
Hidden depth & 3 & 5\\\midrule
Hidden width & 64 & 64\\\midrule
$D_f$ & 128 & ---\\\midrule
$\sigma$ & $3$ & ---\\\midrule
Training iterations & $1\times10^4$ & $3\times10^4$\\\midrule
Initial learning rate & $1\times10^{-3}$ & $5\times10^{-4}$\\\midrule
Learning-rate decay & $0.97$ per $100$ epochs & Cosine to $1\times10^{-6}$\\
\bottomrule
\end{tabular}
\end{table}

\clearpage

\section{Additional experimental results}
\label{app:additional-results}
In this section, we provide additional results on computational
scaling, parameter and topology identification, and trajectory
reconstruction.

\begin{figure}[htbp]
\centering
\includegraphics[width=.8\linewidth]{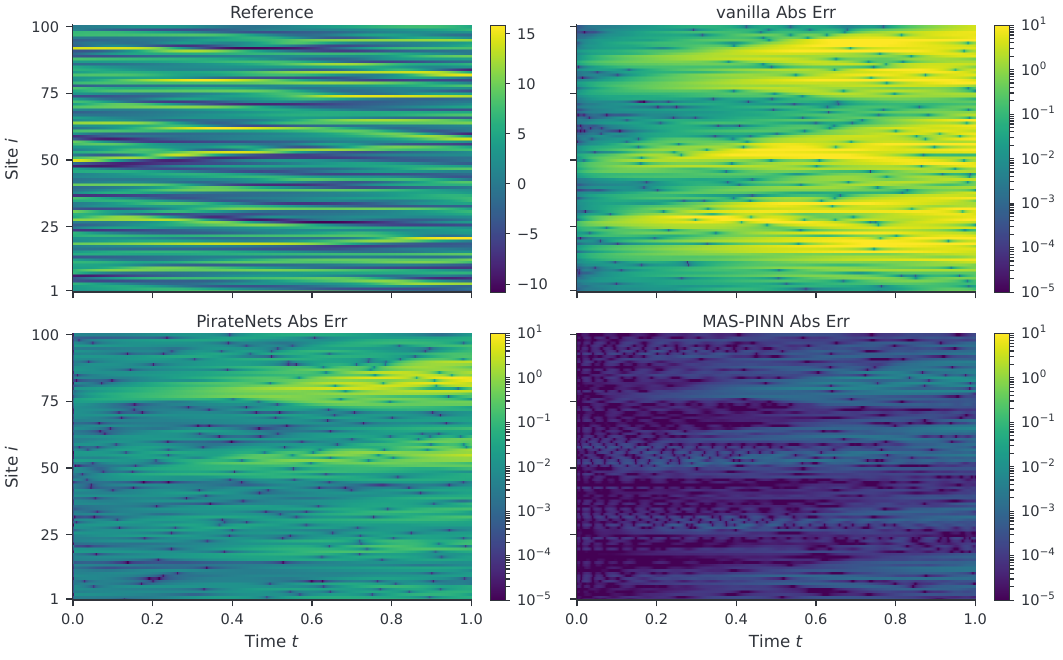}
\caption{Lorenz inference experiment with $N=100$ sites.
Top left: reference solution. Remaining: pointwise absolute
error of the vanilla PINN, PirateNets, and
MAS-PINN on a logarithmic scale.}
\label{fig:lorenz-errors}
\end{figure}

\begin{figure}[htbp]
  \centering
  \includegraphics[width=\linewidth]{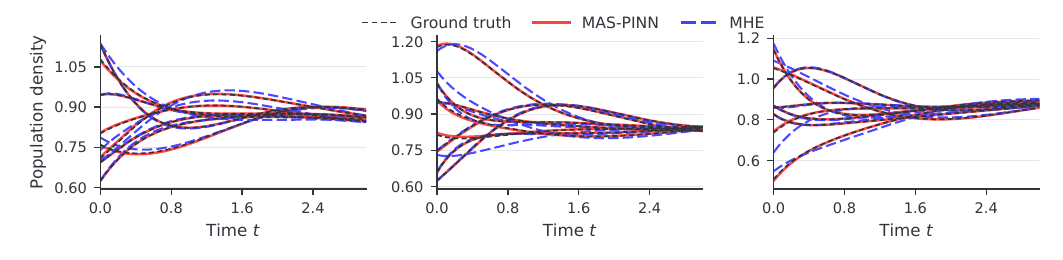}
  \caption{Trajectory prediction across network sites at $0.5\%$ measurement
noise. Left to right: species 1--3. Gray dashed: ground truth;
red solid: MAS-PINN; blue dashed: MHE.}
  \label{fig:lv-trajectories}
\end{figure}

\begin{table}[htbp]
\centering
\caption{One-factor-at-a-time scaling for the partially observed
LV experiment.
Unvaried factors remain at $N=8$, $N_f=1600$, $P=129$,
and 100 CG iterations.
Op.\ denotes one normal-operator application;
LM time covers 100 outer iterations, and memory is the
peak during the LM stage.
``---'' denotes runs omitted due to estimated memory requirements;
``n/a'' denotes not applicable.}
\label{tab:scaling-lv}
\small
\setlength{\tabcolsep}{4pt}
\begin{tabular}{@{}llrrrrr@{}}
\toprule
& & \multicolumn{3}{c}{Matrix-free CG}
  & \multicolumn{2}{c}{Dense LM}\\
\cmidrule(lr){3-5}
\cmidrule(lr){6-7}
Factor & Value
& Op. (ms) & LM (s) & Mem. (MB)
& LM (s) & Mem. (MB)\\
\midrule
$N$ & 4
& 0.35 & 3.63 & 2.8
& 10.9 & 828\\
& 8
& 0.33 & 3.64 & 5.7
& 77.8 & 3169\\
& 16
& 0.45 & 5.06 & 11.6
& 594.0 & 12387\\
& 32
& 0.58 & 6.38 & 23.6
& --- & ---\\
& 64
& 1.36 & 14.39 & 59.5
& --- & ---\\
& 128
& 1.76 & 18.82 & 98.0
& --- & ---\\
& 256
& 3.92 & 41.35 & 216.1
& --- & ---\\
\midrule
$N_f$ & 800
& 0.33 & 3.63 & 2.6
& 45.4 & 1683\\
& 1600
& 0.33 & 3.64 & 5.7
& 77.8 & 3169\\
& 6400
& 0.55 & 5.93 & 19.9
& 271.3 & 12085\\
\midrule
$P$ & 65
& 0.33 & 3.51 & 5.6
& 19.9 & 1577\\
& 129
& 0.33 & 3.64 & 5.7
& 77.8 & 3169\\
& 513
& 0.46 & 5.01 & 9.1
& 1446.9 & 13618\\
\midrule
CG iterations & 30
& 0.33 & 1.20 & 5.7
& n/a & n/a\\
& 60
& 0.33 & 2.25 & 5.7
& n/a & n/a\\
& 100
& 0.33 & 3.64 & 5.7
& n/a & n/a\\
\bottomrule
\end{tabular}
\end{table}

\begin{figure}[htbp]
\centering
\includegraphics[width=\linewidth]{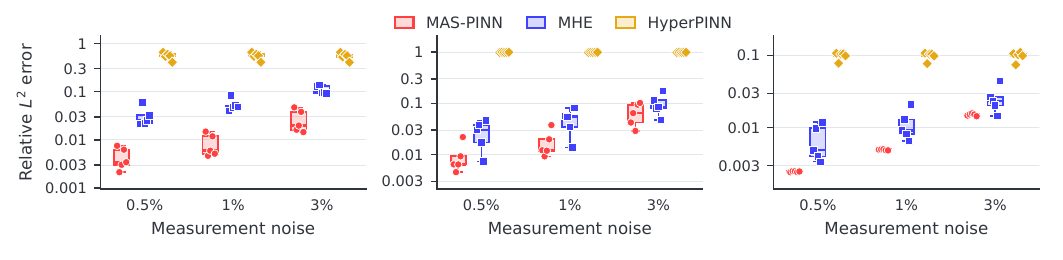}
\caption{Relative $L^2$ error versus measurement noise
for all methods in the Rössler hypergraph experiment.
Left: pairwise weights $w$; middle: three-body weights $v$;
right: reconstructed trajectories.
Boxes show interquartile ranges and medians; whiskers
reach the extreme values within $1.5$ interquartile ranges
of the box. Points show all five trials.}
\label{fig:rossler-errors}
\end{figure}

\begin{figure}[htbp]
\centering
\includegraphics[width=\linewidth]{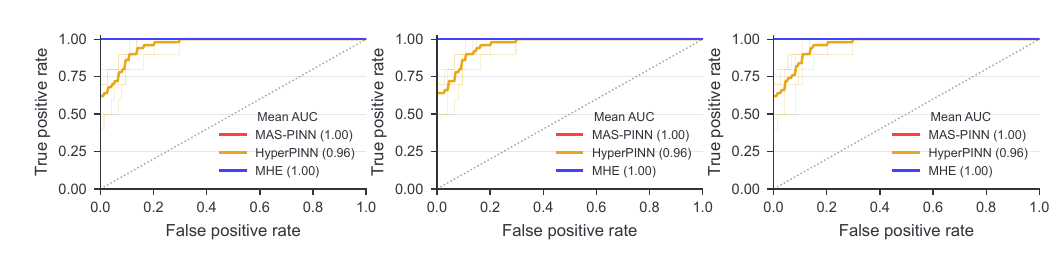}
\caption{ROC curves over all candidate hyperedges
for the Rössler hypergraph experiment.
Left to right: $0.5\%$, $1\%$, and $3\%$ measurement noise.
Thin lines show the five individual trials,
thick lines show their mean, and
legends report mean AUC.}
\label{fig:rossler-roc}
\end{figure}

\begin{figure}[ht]
\centering
\includegraphics[width=\linewidth]{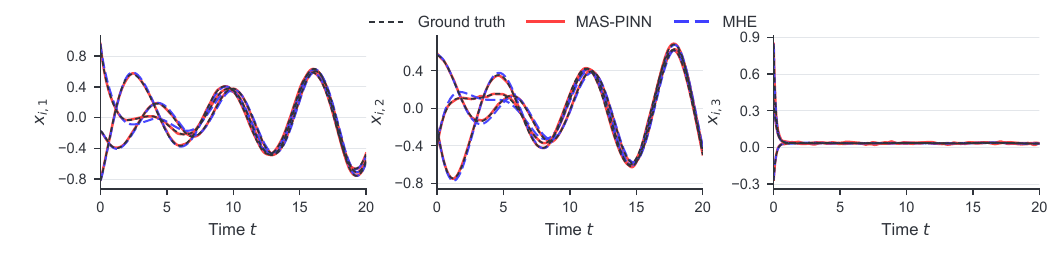}
\caption{Trajectory reconstruction for the Rössler
hypergraph experiment at $3\%$ measurement noise.
Left to right: state components $x_{i,1}$, $x_{i,2}$,
and $x_{i,3}$.
Each panel shows oscillators $2$, $3$, and $4$.
Gray dashed: ground truth; red solid: MAS-PINN;
blue dashed: MHE.}
\label{fig:rossler-trajectories}
\end{figure}

\end{document}